\pdfoutput=1

\documentclass[11pt]{article}

\usepackage{acl}

\usepackage{times}
\usepackage{latexsym}
\usepackage{booktabs}
\usepackage{colortbl}
\usepackage{pythonhighlight}
\usepackage{xspace}
\usepackage{graphicx}
\usepackage{pifont}
\definecolor{lp}{HTML}{CBC3E3}

\usepackage[T1]{fontenc}

\usepackage[utf8]{inputenc}

\usepackage{microtype}

\usepackage{inconsolata}

\usepackage{graphicx}

\usepackage{multirow}
\usepackage{listings}
\usepackage{amssymb}
\usepackage{amsmath}

\newcommand\blfootnote[1]{%
  \begingroup
  \renewcommand\thefootnote{}\footnote{#1}%
  \addtocounter{footnote}{-1}%
  \endgroup
}

\title{Understanding LLM Development Through Longitudinal Study: \\Insights from the Open Ko-LLM Leaderboard}

\author{Chanjun Park$^{1\dagger}$, Hyeonwoo Kim$^{2}$ \\
\\
 $^{1}$Korea University, $^{2}$Upstage AI \\
  \texttt{bcj1210@korea.ac.kr} \\
  \texttt{choco\_9966@upstage.ai}}

\begin{document}
\maketitle
\begin{abstract}
\blfootnote{$^\dagger$ Corresponding Author}
This paper conducts a longitudinal study over eleven months to address the limitations of prior research on the Open Ko-LLM Leaderboard, which have relied on empirical studies with restricted observation periods of only five months. By extending the analysis duration, we aim to provide a more comprehensive understanding of the progression in developing Korean large language models (LLMs). Our study is guided by three primary research questions: (1) What are the specific challenges in improving LLM performance across diverse tasks on the Open Ko-LLM Leaderboard over time? (2) How does model size impact task performance correlations across various benchmarks? (3) How have the patterns in leaderboard rankings shifted over time on the Open Ko-LLM Leaderboard?. By analyzing 1,769 models over this period, our research offers a comprehensive examination of the ongoing advancements in LLMs and the evolving nature of evaluation frameworks.
\end{abstract}

\section{Introduction}
The rapid advancement of large language models (LLMs)~\cite{zhao2023survey} has led to the creation of various leaderboards designed to evaluate their performance across a wide range of tasks~\cite{alpaca_eval,lee2023holistic,HughesBae2023,bigcodeleaderboard,li2023halueval}. Among these, the Open LLM Leaderboard~\cite{open-llm-leaderboard-v1,open-llm-leaderboard-v2} developed by Hugging Face~\cite{jain2022hugging} has achieved significant global recognition. In the context of Korean language models, the Open Ko-LLM Leaderboard~\cite{park2024open} was established to specifically assess LLM performance within the Korean language environment.

While previous analyses of the Open Ko-LLM Leaderboard~\cite{park2024open} have provided valuable insights into LLM performance, they have been constrained observation periods of only five months, limiting their ability to capture long-term trends. To better understand the ongoing evolution and inherent challenges in LLM development, a more comprehensive and extended analysis is required. This paper addresses this gap by conducting a detailed longitudinal study of the Open Ko-LLM Leaderboard, guided by three primary research questions:

\textbf{First}, we analyze the longitudinal changes in performance across five tasks monitored by the Open Ko-LLM Leaderboard. These tasks are designed to evaluate various capabilities of LLMs, including reasoning, natural language understanding, and common sense knowledge. By examining data collected over a eleven-month period, this study aims to identify which capabilities have presented the greatest challenges for LLM developers, which tasks have reached performance saturation rapidly, and which tasks continue to pose significant difficulties. This analysis will provide quantitative insights into performance trends across different tasks, thereby guiding targeted research efforts and highlighting key areas that require further advancement to push the boundaries of model development.

\textbf{Second}, we explore the correlations between different tasks based on model size. This aspect of the study examines how the performance across different tasks varies depending on the scale of the model. Understanding these correlations will provide insights into the interaction between model capacity and task performance, offering a deeper understanding of how scaling influences overall effectiveness across tasks.

\textbf{Third}, we examine the evolution of leaderboard dynamics from the initial stages to the present by focusing on three key aspects: the correlations between task performances in the early months compared to the entire eleven-month period, the temporal changes in performance based on model type, and the shifts in performance relative to model size. This comprehensive analysis offers insights into the evolving interplay among tasks and the influence of various model characteristics on LLM performance throughout different phases of development.

\begin{figure}[t!]
    \centering
    \resizebox{1.00\linewidth}{!}{
\includegraphics{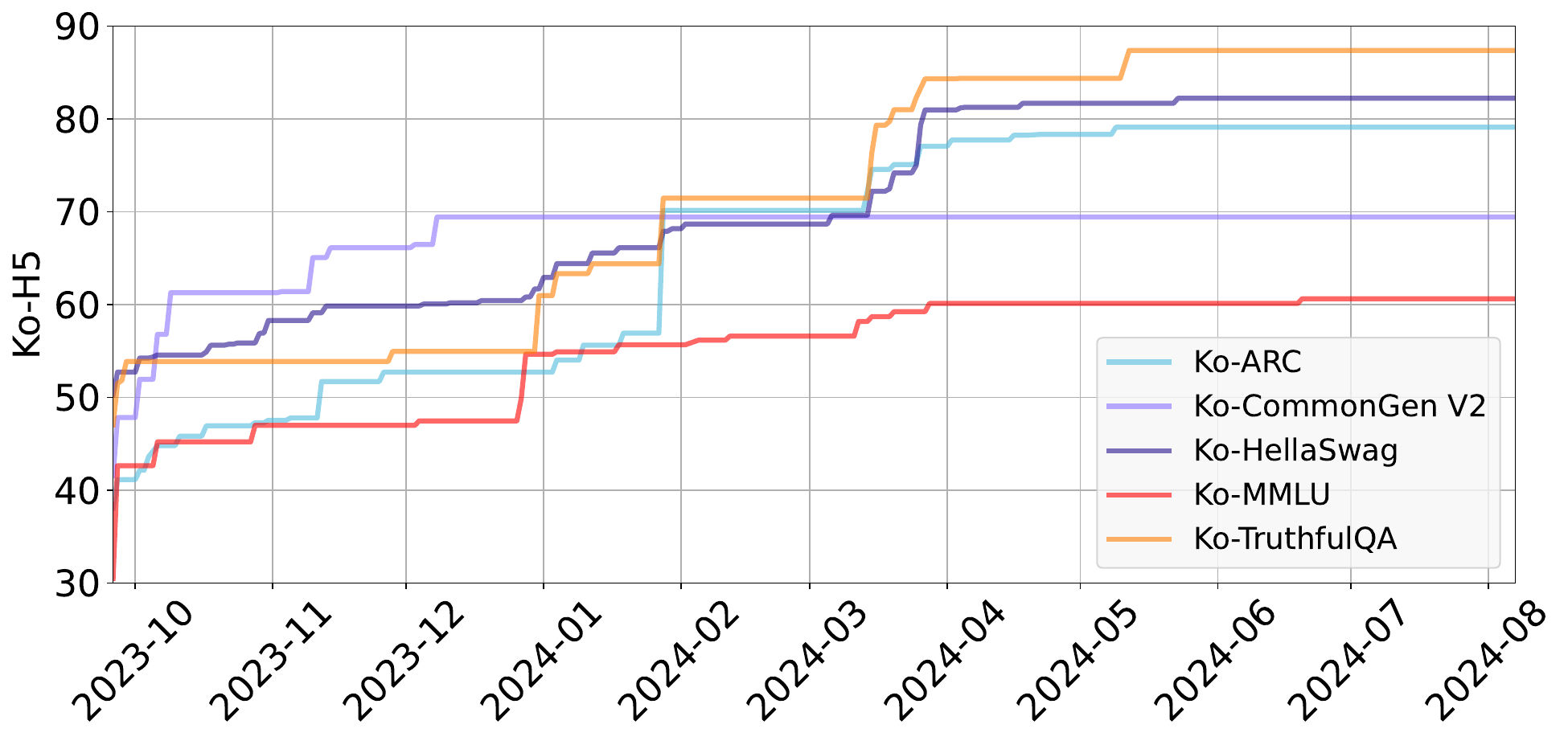}
    }
    \caption{Performance trends of LLMs across different tasks on the Open Ko-LLM Leaderboard over a eleven-month period. The total number of submitted models is 1,769.}
    \label{fig:koh5_each_metric}
\end{figure}

\section{Open Ko-LLM Leaderboard}
The Open Ko-LLM Leaderboard~\cite{park2024open} is a pioneering platform designed to evaluate large language models (LLMs) specifically in the Korean language, addressing the limitations of predominantly English-focused benchmarks. This leaderboard mirrors the structure of the globally recognized Open LLM Leaderboard by Hugging Face~\cite{open-llm-leaderboard-v1}, ensuring consistency and comparability across languages. It is built on two key principles: alignment with the English leaderboard and the use of private test sets to avoid data contamination, thereby enhancing evaluation robustness.

The leaderboard employs the Ko-H5 benchmark, comprising five tasks that assess various aspects of language understanding and generation in Korean. These tasks are designed to comprehensively evaluate LLM capabilities. The first task, Ko-Hellaswag~\cite{zellers2019hellaswag}, tests commonsense reasoning by requiring models to complete sentences contextually and logically. The second task, Ko-ARC~\cite{clark2018think}, adapted from the English ARC, evaluates both commonsense and scientific reasoning through multiple-choice questions. Ko-MMLU~\cite{hendrycks2020measuring}, the third task, assesses multitask language understanding and domain knowledge across various subjects, requiring models to respond accurately to questions from different domains. The fourth task, Ko-CommonGen V2~\cite{seo2024kocommongen}, focuses on commonsense generation, where models must create coherent sentences from given concepts, testing their ability to connect common knowledge meaningfully. Lastly, Ko-TruthfulQA~\cite{lin2021truthfulqa} evaluates a model ability to provide truthful and accurate responses, crucial for assessing the factual integrity of LLMs in real-world scenarios.

Through the Ko-H5 benchmark, the Open Ko-LLM Leaderboard provides a robust framework for evaluating Korean LLMs and promotes linguistic diversity in LLM evaluation. By incorporating tasks that reflect Korean linguistic and cultural nuances, the leaderboard offers valuable insights into LLM performance beyond English, encouraging a more inclusive approach to language model evaluation.

\section{Empirical Analysis}
\label{sec:empirical}
\subsection{Challenges in Enhancing Task Performance Over Time}
\textit{What are the specific challenges in improving LLM performance across diverse tasks on the Open Ko-LLM Leaderboard over time?}. To investigate this question, we conducted a comprehensive analysis of performance trends over a eleven-month period across all tasks on the Open Ko-LLM Leaderboard, including Ko-HellaSwag (commonsense reasoning)\cite{zellers2019hellaswag}, Ko-ARC (commonsense and scientific reasoning)\cite{clark2018think}, Ko-MMLU (multitask language understanding and domain knowledge)\cite{hendrycks2020measuring}, Ko-CommonGEN V2 (commonsense generation)\cite{seo2024kocommongen}, and TruthfulQA (truthfulness)~\cite{lin2021truthfulqa}.

Figure~\ref{fig:koh5_each_metric} and Table~\ref{tab:weeks_to_scores} show the varying performance patterns of LLMs across these tasks over the eleven-month period. Certain tasks, such as Ko-HellaSwag and Ko-TruthfulQA, exhibit rapid improvements in performance and early saturation. Specifically, Ko-HellaSwag reached a score of 50 almost immediately and achieved 80 by week 26, while Ko-TruthfulQA showed comparable progress, reaching a score of 80 within 25 weeks. These trends indicate that current LLMs are particularly well-suited for tasks requiring straightforward commonsense reasoning and truthfulness, suggesting a relatively lower barrier to achieving performance enhancements in these domains.

Conversely, tasks such as Ko-MMLU and Ko-CommonGEN V2 show slower, more gradual improvements without clear signs of saturation, highlighting their increased complexity and the deeper understanding required from LLMs. Ko-MMLU took 13 weeks to reach a score of 50 and then stabilized around 60 after 26 weeks, indicating a limit to the current models capabilities. Similarly, Ko-CommonGEN V2, despite reaching a score of 50 relatively quickly, showed minimal progress beyond 60. These patterns highlight the significant challenges LLMs face in tasks that demand complex reasoning and specialized knowledge, suggesting these are important areas for further research.

The initial rapid gains in Ko-ARC, followed by minimal progress beyond a score of 60 after 17 weeks, indicate that while LLMs can quickly adapt to certain tasks, their progress is constrained by the need for more complex reasoning skills. This underscores the importance of developing more challenging benchmarks to better evaluate the limitations and capabilities of LLMs, especially in tasks that require more advanced forms of reasoning.

Overall, these findings emphasize the need to include a broad range of complex tasks to comprehensively assess LLM capabilities. While some tasks demonstrate rapid performance saturation, others present ongoing challenges, serving as essential benchmarks for guiding future advancements in LLM development.

\begin{table}[t!]
    \centering
    \resizebox{1.0\linewidth}{!}{
        \begin{tabular}{lcccc}
        \toprule
        Dataset & 50 & 60 & 70 & 80 \\
        \hline
        Ko-ARC & $\sim$ 6 & $\sim$ 17 & $\sim$ 17 & - \\
        Ko-HellaSwag & $\sim$ 0 & $\sim$ 10 & $\sim$ 24 & $\sim$ 26 \\
        Ko-MMLU & $\sim$ 13 & $\sim$ 26 & - & - \\
        Ko-TruthfulQA & $\sim$ 0 & $\sim$ 13 & $\sim$ 17 & $\sim$ 25 \\
        Ko-CommonGen V2 & $\sim$ 0 & $\sim$ 1 & - & - \\
        \bottomrule
        \end{tabular}
    }
    \caption{Number of weeks it took to reach scores of 50, 60, 70, and 80 out of 100 for the individual tasks.}
    \label{tab:weeks_to_scores}
\end{table}

\begin{figure*}[t!]
    \centering
    \resizebox{0.90\linewidth}{!}{
    \includegraphics{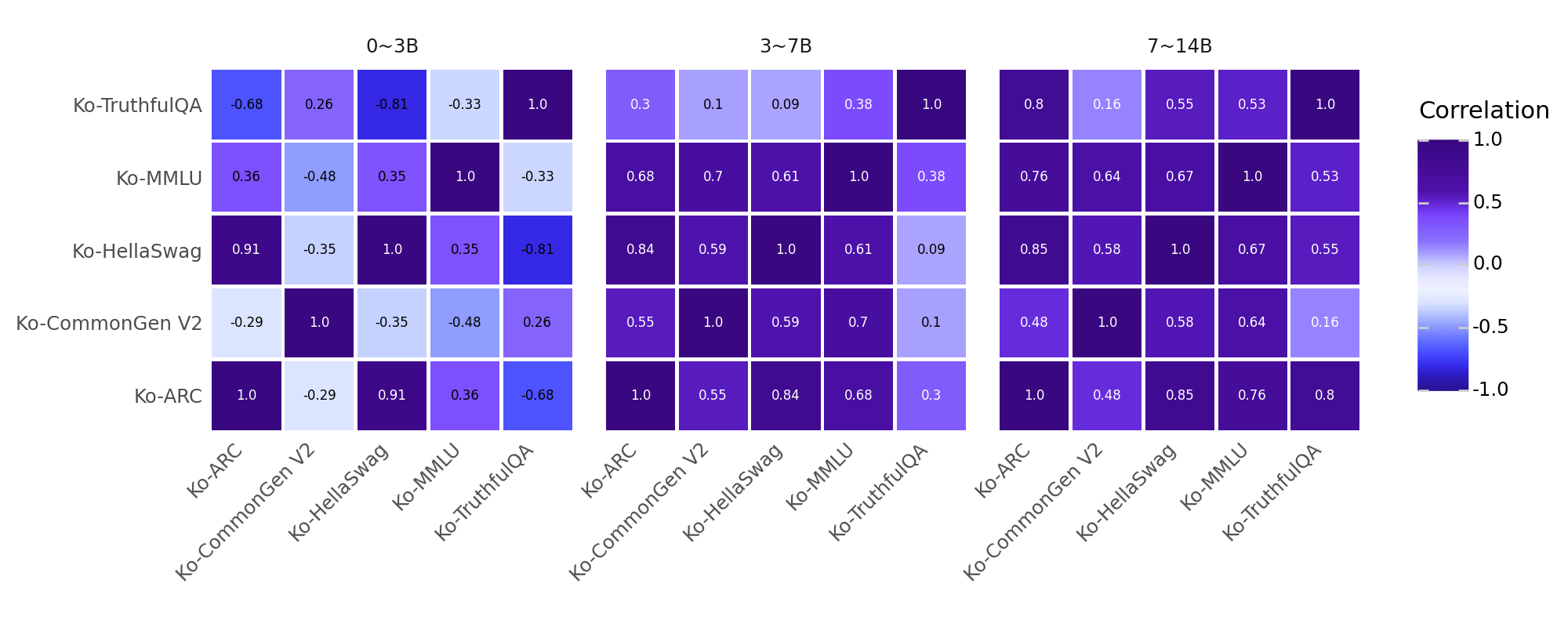}
    }
    \caption{Correlation between task performances across different model size categories, illustrating how task correlations change with increasing model size.}
    \label{fig:param_corr}
\end{figure*}

\subsection{The Influence of Model Size on Task Performance Correlations}
\textit{How does model size impact task performance correlations across various benchmarks?}. To investigate this question, we analyze how model size affects performance improvements across different tasks, using a framework similar to previous studies~\cite{park2024open}. For this analysis, models were divided into three size categories: under 3 billion parameters, 3 to 7 billion parameters, and 7 to 14 billion parameters. This categorization allows for a detailed examination of how scaling impacts task performance.

Figure~\ref{fig:param_corr} illustrates distinct patterns in task performance correlations depending on model size. Smaller models (under 3 billion parameters) show low or even negative correlations between certain tasks, such as Ko-TruthfulQA and Ko-CommonGen V2, and other tasks. This suggests that smaller models struggle to improve consistently across multiple capabilities, indicating that advancements in one area do not necessarily lead to improvements in others. Consequently, these models tend to have a fragmented skill set, making them less suitable for a comprehensive evaluation of LLM performance.

In contrast, larger models demonstrate higher correlations across most tasks, suggesting that increasing model size results in a more effective integration of various capabilities. For example, models in the 7 to 14 billion parameter category exhibit stronger positive correlations across a majority of tasks, especially those requiring advanced reasoning. This trend indicates that scaling up model size not only enhances performance on individual tasks but also supports a more cohesive development of capabilities, enabling more consistent performance improvements across a wide range of tasks.

These findings highlight the importance of model size in achieving balanced performance across a range of tasks. Smaller models, with their inconsistent performance across tasks, suggest a limitation in their ability to generalize learning effectively. In contrast, the positive correlations observed in larger models imply that increasing model size fosters a more comprehensive understanding and transfer of knowledge across different domains. This insight is crucial for future LLM development, as it underscores the need to consider model size not just for boosting individual task performance, but also for promoting a more integrated and holistic enhancement of capabilities.

\subsection{Temporal Shifts in Leaderboard Ranking Patterns}
\label{sec:3.3}
\textit{How have the patterns in leaderboard rankings shifted over time on the Open Ko-LLM Leaderboard?}. To investigate this question, we extended our analysis to an eleven-month period to see if the initial trends, defined as those observed during the initial five months in the previous study by~\citet{park2024open}, remained consistent or if new patterns emerged over time. This longer timeframe allows us to capture shifts in model performance and ranking dynamics.

\begin{figure*}[t!]
    \centering
    \resizebox{0.65\linewidth}{!}{
    \includegraphics{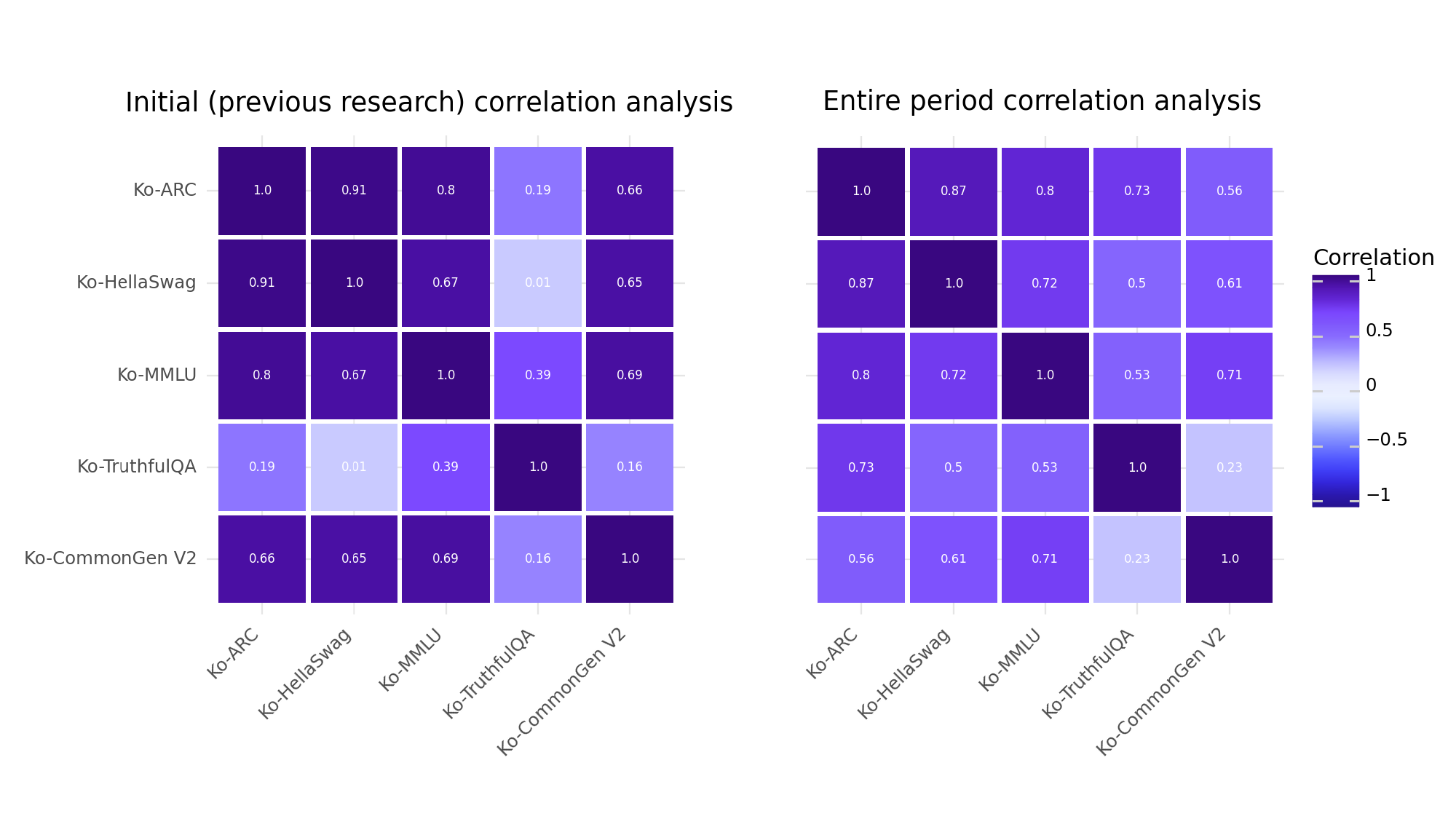}
    }
    \caption{Analysis of Task Correlations Over Time.}
    \label{fig:correlation_old}
\end{figure*}

\begin{figure}[t!]
    \centering
    \resizebox{0.85\linewidth}{!}{
    \includegraphics{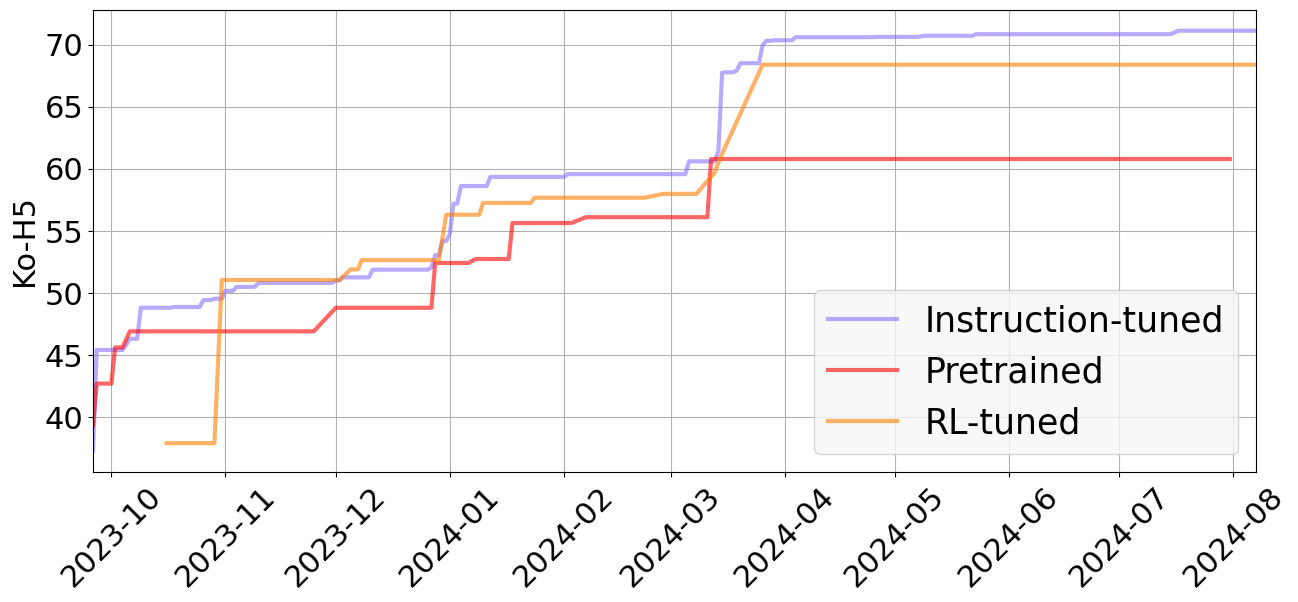}
    }
    \caption{Performance Trends Over Time for Different Model Types.}
    \label{fig:per_type}
\end{figure}

\begin{figure}[t!]
    \centering
    \resizebox{0.85\linewidth}{!}{
    \includegraphics{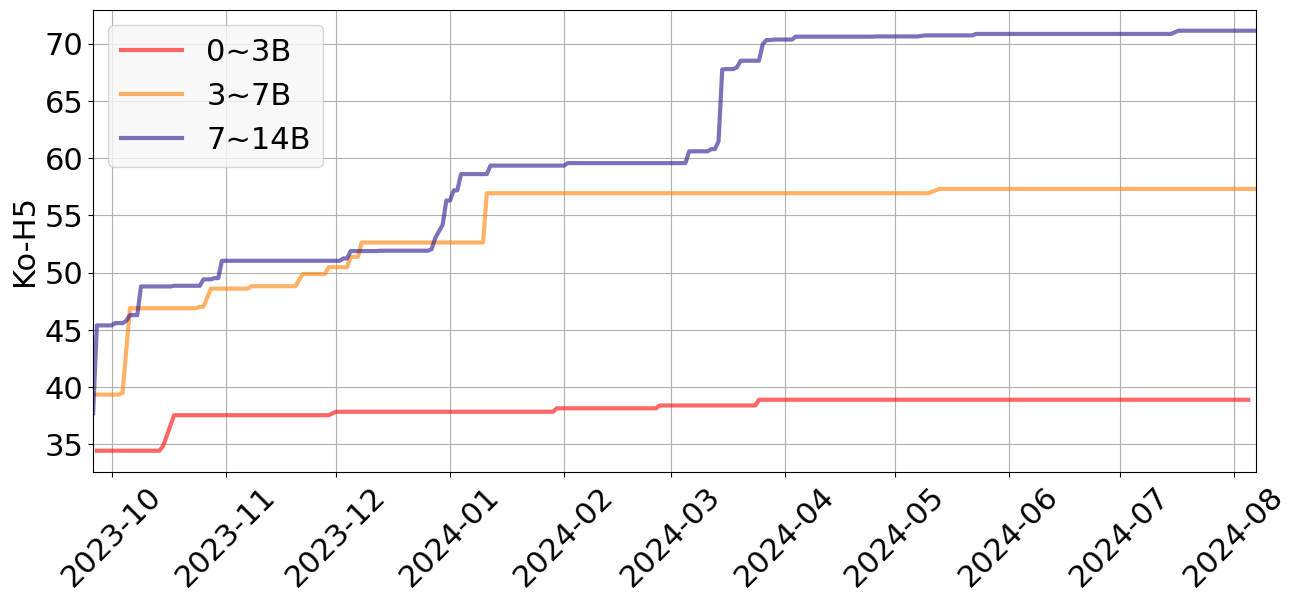}
    }
    \caption{Performance Trends by Model Size.}
    \label{fig:per_size}
\end{figure}

\begin{figure*}[t!]
    \resizebox{1.0\linewidth}{!}{
    \includegraphics{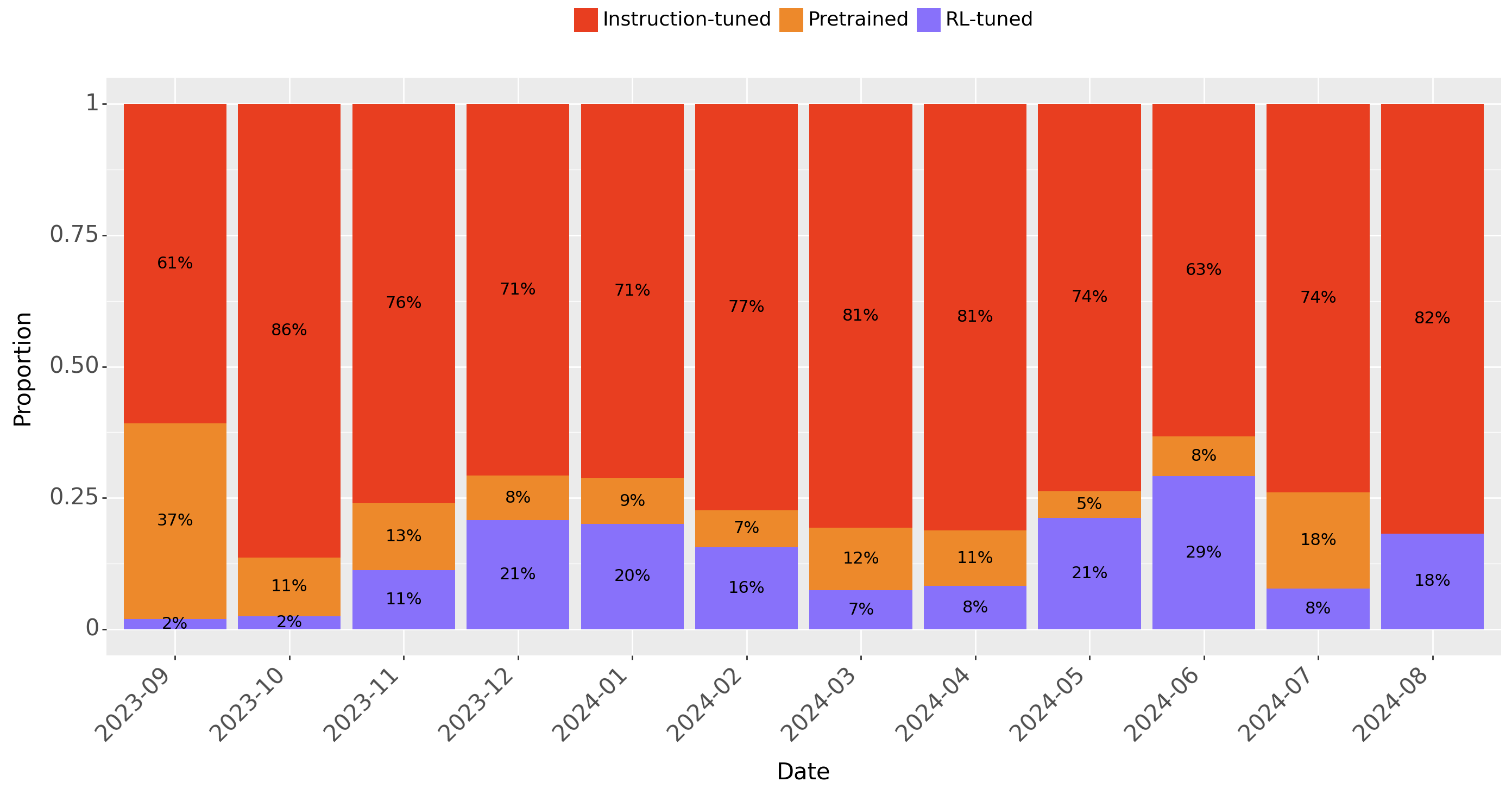}
    }
    \caption{Monthly distribution of submissions by model type on the Open Ko-LLM leaderboard.}
    \label{fig:monthly_proportion}
\end{figure*}

\begin{table*}[h]
\centering
\begin{tabular}{|l|c|c|}
\hline
\textbf{Date} & \textbf{Model Submissions Count} & \textbf{Model Evaluation Count} \\ \hline
2023-09       & 51                         & 40                        \\ 
2023-10       & 322                        & 255                       \\ 
2023-11       & 337                        & 280                       \\ 
2023-12       & 260                        & 225                       \\ 
2024-01       & 289                        & 234                       \\ 
2024-02       & 115                        & 99                        \\ 
2024-03       & 176                        & 153                       \\ 
2024-04       & 170                        & 122                       \\ 
2024-05       & 156                        & 134                       \\ 
2024-06       & 79                         & 72                        \\ 
2024-07       & 142                        & 129                       \\ 
2024-08       & 33                         & 26                        \\ \hline
\textbf{Total} & \textbf{2230}              & \textbf{1769}             \\ \hline
\end{tabular}
\caption{Monthly distribution of model submissions and evaluation on the Open Ko-LLM leaderboard.}
\label{tab:monthly_submissions_model_runs}
\end{table*}

\paragraph{Task Correlations Over Time.} Figure~\ref{fig:correlation_old} shows the correlation analysis between tasks during the initial phases of the leaderboard and over the full eleven-month period. A notable increase was observed in the correlation between Ko-Truthful QA and other tasks, especially Ko-Hellaswag. This correlation, initially very low at 0.01, rose significantly to 0.5 over time. This change suggests that as higher-performing models, particularly those with 7 billion parameters or more, were introduced, the alignment between tasks became stronger. For most other tasks, correlations remained relatively stable, reflecting their initial patterns.

\paragraph{Performance Trends by Model Type.} Figure~\ref{fig:per_type} presents the performance trends over time for different model types. As noted in previous research~\cite{park2024open}, improvements in instruction-tuned models typically lagged behind those of pretrained models by about one week. When a pretrained model showed a significant performance boost, instruction-tuned models followed with a similar increase roughly one week later. This pattern persisted throughout the entire period analyzed, indicating a reliance of instruction-tuned models on the advancements made by pretrained models. After April 2024, the performance of pretrained models stabilized, leading to a corresponding lack of progress in both instruction-tuned and RL-tuned models. This trend indicates the fundamental role of pretrained models in driving overall performance gains in LLMs and suggests that further improvements in pretrained models are necessary for advancing model capabilities.

\paragraph{Performance Trends Across Model Sizes.} Figure~\ref{fig:per_size} shows performance variations by model size. Models in the 0-3B range exhibited minimal improvement throughout the leaderboard period, indicating inherent scalability limitations. Similarly, models in the 3-7B range initially demonstrated gains, but their progress stabilized around five months in (April 2024 to August 2024), revealing similar scalability constraints.

Larger models in the 7-14B range showed steady performance improvements during the early phase of the leaderboard, continuing throughout the entire analysis period. However, after April 2024, their performance also reached a saturation point. This stagnation is likely due to the absence of new, high-performing Korean pretrained models, a trend also evident in the analysis of different model types in Figure~\ref{fig:per_type}.

These findings emphasize that improving LLM performance largely depends on advancements in pretrained models. The leaderboard analysis indicates that, without new breakthroughs in pretrained models, further improvements are limited. This highlights the essential role of continuous innovation in pretrained models for advancing LLM performance.

\subsection{Evaluation Patterns and Submission Insights}
Figure~\ref{fig:monthly_proportion} presents the monthly distribution of submissions across different model types on the Open Ko-LLM leaderboard. Initially, pretrained models constituted 37\% of all submissions, but this proportion declined sharply over time, with no pretrained models submitted by August 2024. This trend signals a diminishing focus on pretrained models within the community, which is concerning given their foundational importance discussed in Section~\ref{sec:3.3}. Therefore, a renewed emphasis on fostering interest and engagement with pretrained models could help address this emerging gap.

On the other hand, instruction-tuned models, which started at 61\%, consistently dominated the submissions, maintaining a steady presence of 70-80\% each month. This trend suggests that the community perceives instruction-tuned models as highly effective or suitable for the tasks evaluated. Additionally, RL-tuned models, though initially making up only 2\% of submissions, gradually increased to a peak of 29\%, reflecting a growing interest in exploring reinforcement learning approaches within the leaderboard context. This variety indicates a healthy exploration of diverse model types, but also highlights areas where community focus could be broadened or rebalanced.

In addition, Table~\ref{tab:monthly_submissions_model_runs} presents the monthly statistics for both the number of model submissions and the number of completed model evaluations. The \textit{Model Submissions Count} refers to the total number of models submitted to the leaderboard each month. In contrast, the \textit{Model Evaluation Count} represents the number of these submitted models that successfully completed the evaluation process.

The discrepancy between the \textit{Model Submissions Count} and the \textit{Model Evaluation Count} is due to instances where some models fail to complete the evaluation phase on the leaderboard. This failure can occur for several reasons, such as models being too large to be processed within the available computational resources or issues related to library support and compatibility. As a result, not all submitted models are evaluated successfully, highlighting potential challenges and areas for improvement in handling diverse model architectures on the leaderboard.

\section{Conclusion}
This study provides a longitudinal analysis of the Open Ko-LLM Leaderboard, uncovering significant performance trends and underlying challenges in LLM development. It was observed that smaller models consistently face scalability limitations, preventing substantial performance advancements. In contrast, larger models initially show promising improvements but eventually reach a saturation point, highlighting a critical dependency on advancements in pretrained models. These findings underscore the need for continuous innovation and enhancement in the development of pretrained models to push the boundaries of LLM capabilities further. Additionally, the analysis demonstrates the utility of leaderboard data in tracking the evolving dynamics of LLM performance. By examining a broader range of model submissions and evaluation patterns over an extended period, this study provides valuable insights into how model size, type, and tuning methods influence overall effectiveness. Such insights can inform targeted research efforts and encourage the development of strategies aimed at overcoming existing limitations, ultimately supporting more robust and adaptable LLMs.

\section*{Acknowledgments}
We sincerely thank the National Information Society Agency (NIA), Korea Telecom (KT), and Flitto for their support. We also extend our gratitude to the Korea University NLP \& AI Lab, particularly Professor Heuiseok Lim and Jaehyung Seo, for their valuable data contributions, which have greatly enhanced the robustness of the leaderboard. Our appreciation goes to the Hugging Face teams, especially Clémentine Fourrier, Lewis Tunstall, Omar Sanseviero, and Philipp Schmid, for their assistance. We would like to thank SeongHwan Cho for his contributions to the leaderboard development, and Sanghoon Kim for his contributions to the leaderboard infrastructure. Special thanks to Hyunbyung Park for his initial contributions to Ko-H5. 

We are also grateful to Professor Harksoo Kim from Konkuk University, Professor Hwanjo Yu from Pohang University of Science and Technology, Professor Sangkeun Jung from Chungnam National University, and Professor Alice Oh from KAIST for their insightful advice on the Open Ko-LLM Leaderboard. Finally, we deeply appreciate the open-source community for their invaluable feedback and contributions.

This work was supported by Institute of Information \& Communications Technology Planning \& Evaluation(IITP) grant funded by the Korea government(MSIT) (No. RS-2024-00338140, Development of learning and utilization technology to reflect sustainability of generative language models and up-to-dateness over time).

\section*{Limitations}
While this study provides valuable insights into the evaluation of LLMs, several limitations should be acknowledged. First, our analysis is primarily based on data from the Open Ko-LLM Leaderboard. Although this leaderboard offers extensive coverage of various tasks, it may not fully represent the complete spectrum of challenges and scenarios relevant to LLM performance, particularly in specialized or emerging domains. 

Additionally, the focus on Korean language models may restrict the generalizability of our findings to other languages and cultural contexts. The linguistic and cultural nuances specific to Korean may not entirely translate to other languages, potentially limiting the applicability of our conclusions. 

Furthermore, our study predominantly examines the relationship between model size and performance but does not explore other factors, such as training data diversity or the impact of different fine-tuning techniques, which could also significantly influence model outcomes. Future research should aim to address these gaps by incorporating a broader range of tasks, languages, and evaluation metrics. Expanding the scope of analysis to include models trained in different linguistic and cultural settings, as well as exploring the impact of varied training methodologies, would enhance the robustness and applicability of the findings.

\section*{Ethics Statement}
In conducting this research, we adhered to the highest ethical standards, ensuring that all data used in the evaluation was sourced responsibly and in compliance with relevant regulations. We are committed to transparency and integrity in our research practices, and we have made our methods and findings available to the community for further scrutiny and development. We also acknowledge the importance of considering the societal impacts of LLMs, particularly in ensuring that their development and deployment are aligned with ethical principles that promote fairness, inclusivity, and accountability.

\bibliography{custom}

\end{document}